\documentclass[10pt,twocolumn,letterpaper]{article}

\usepackage{cvpr}              
\definecolor{cvprblue}{rgb}{0.21,0.49,0.74}
\usepackage[pagebackref,breaklinks,colorlinks,allcolors=cvprblue]{hyperref}

\usepackage{multirow}
\usepackage{lipsum} 
\usepackage{soul}
\usepackage{adjustbox}

\def\paperID{*****} 
\def\confName{CVPR}
\def\confYear{2026}

\title{GlowGS: Generative Semantic Feature Learning for \\ 3D Gaussian Splatting in Nighttime Glow Scenes}

\author{
    Beibei Lin \qquad
    Xiao Cao \qquad
    Jingyuan Guo \qquad
    Robby T. Tan
    \\
    National University of Singapore\\
    {\tt\small beibei.lin@u.nus.edu, robby.tan@nus.edu.sg}
    }

\begin{document}
\maketitle

\begin{abstract}
Existing 3DGS methods effectively render high-quality novel views in clear-day scenes. However, they struggle with night scenes, particularly in glow regions, due to the lack of structural features such as textures and edges, which are key cues for splatting-based reconstruction. To address this problem, we leverage a diffusion model and a Vision Foundation Model (VFM) to compensate for missing structural cues. Our method consists of two key novel ideas: semantic feature generation and novel-view semantic learning. First, semantic feature generation produces high-quality semantic features as implicit structural cues for novel views. Specifically, a diffusion model synthesizes novel views with unknown camera poses from training views, while a VFM evaluates their quality. Once high-quality novel views are identified, the VFM extracts robust features to construct the semantic feature bank. Second, novel-view semantic learning enables 3DGS to optimize rendered novel views without requiring ground truth. It achieves this by extracting semantic features from a rendered novel view, searching the feature bank for the most similar features, and minimizing their distance. This process enforces implicit structural constraints, ensuring semantically coherent, artifact-free rendered views. Extensive experiments demonstrate the effectiveness of our GlowGS in generating semantically accurate 3D views, showing significant  improvements over existing methods.  
\end{abstract}

\section{Introduction}
\label{sec:intro}
\begin{figure*}
  \centering
  \begin{subfigure}{0.18\linewidth}
    \includegraphics[width=\linewidth]{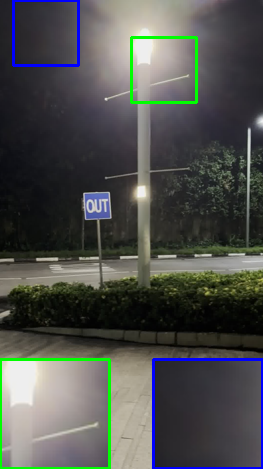}
    \caption{GT}
    \label{fig:short-a}
  \end{subfigure}
  \hfill
  \begin{subfigure}{0.18\linewidth}
    \includegraphics[width=\linewidth]{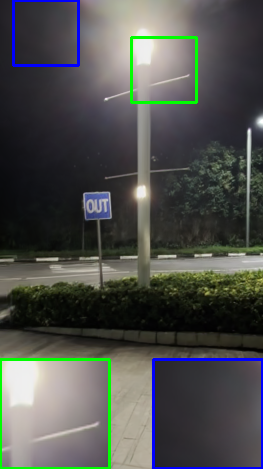}
    \caption{Ours}
    \label{fig:short-e}
  \end{subfigure}
  \hfill
  \begin{subfigure}{0.18\linewidth}
    \includegraphics[width=\linewidth]{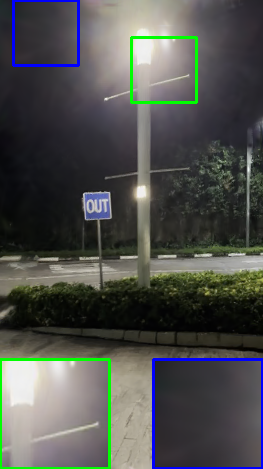}
    \caption{3DGS}
    \label{fig:short-b}
  \end{subfigure}
  \hfill
  \begin{subfigure}{0.18\linewidth}
    \includegraphics[width=\linewidth]{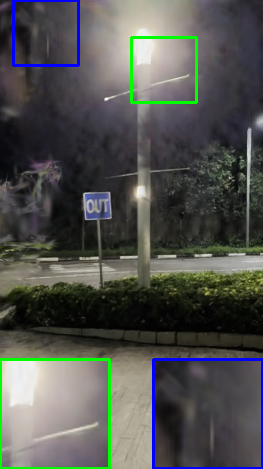}
    \caption{CGS}
    \label{fig:short-c}
  \end{subfigure}
  \hfill
  \begin{subfigure}{0.18\linewidth}
    \includegraphics[width=\linewidth]{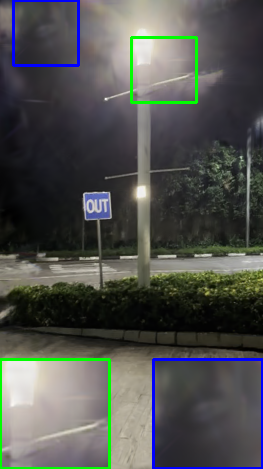}
    \caption{MGS}
    \label{fig:short-d}
  \end{subfigure}
  \caption{
  Qualitative results from 3DGS \cite{kerbl20233d}, CGS \cite{lee2023compact}, MGS \cite{yu2023mip}, and our method on nighttime scenes. All results are from novel views. 
  The zoom-in insets highlight artifacts in glow regions produced by existing methods, including duplicated light sources and unnatural bright spots in dark areas. 
  Our method reconstructs glow regions with smoother light diffusion, effectively reducing these artifacts. 
  }
  \label{Fig_trailor}
\end{figure*}

Existing 3D Gaussian Splatting (3DGS) methods (e.g.,~\cite{kerbl20233d, yu2023mip, lee2023compact}) demonstrate strong performance in rendering novel views of clear-day scenes but struggle with complex nighttime scenes, particularly in glow regions that lack distinct structural features \cite{jin2023enhancing, lin2025nighthaze}. 
These regions exhibit intensity and color variations without physical surfaces, preventing 3DGS from capturing reliable geometric structures.
While 3DGS effectively models these areas in training views, it fails to generalize, leading to artifacts in novel view synthesis. As shown in Figure~\ref{Fig_trailor}, existing 3DGS methods perform poorly in glow regions.

In this paper, we introduce \textbf{GlowGS}, a nighttime 3DGS method for rendering novel views of night scenes with prominent glow regions. GlowGS is built on two key components: semantic feature generation and novel-view semantic learning.
Our semantic feature generation employs a diffusion model and a Vision Foundation Model (VFM), such as DINO \cite{caron2021emerging} and CLIP \cite{radford2021learning},  to produce high-quality semantic features, providing implicit structural cues for novel views. As shown in Figure~\ref{Fig_Dino}, the VFM effectively generates discriminative representations in glow regions.
We use an image-to-video diffusion model, such as Pika~\cite{pika2024} or PromeAI~\cite{promeai2024}, to synthesize novel views with unknown camera poses from training images, enabling the capture of multi-view glow information.

However, the generation process may introduce inaccuracies and artifacts. To address this, we use a VFM-based verification process to assess the quality of generated novel views. The VFM extracts features from both training and generated views, measures their similarity, and selects high-quality novel views accordingly. Once identified, it extracts robust features to construct the semantic feature bank.
Since these features are extracted from generated views with unknown camera poses, they cannot be directly used for novel view training.
To overcome this, we introduce novel-view semantic learning, enabling 3DGS to optimize novel views using the semantic feature bank.
This approach builds on the idea that similar regions share closely related semantic features despite minor pixel variations. As a result, the semantic features in our bank constrain corresponding regions in novel views rendered by 3DGS.

\begin{figure}
  \centering

    \begin{subfigure}{0.32\linewidth}
    \includegraphics[width=\linewidth]{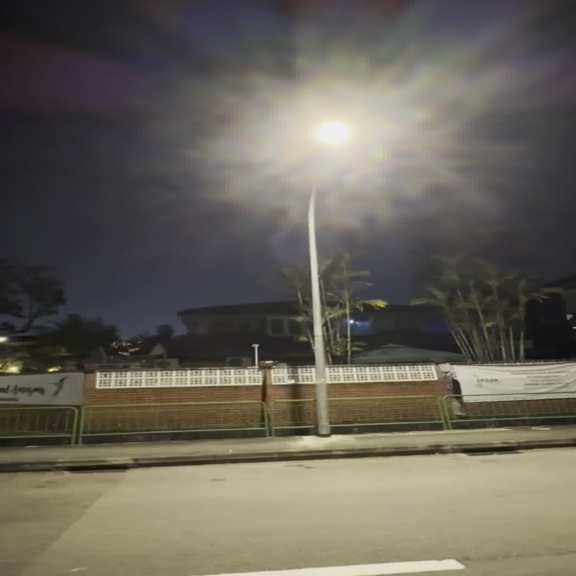}
    \end{subfigure}
    \begin{subfigure}{0.32\linewidth}
    \includegraphics[width=\linewidth]{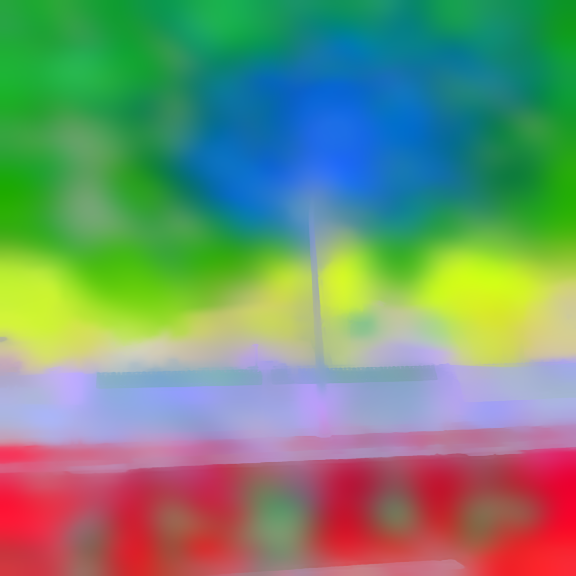}
    \end{subfigure}
    \begin{subfigure}{0.32\linewidth}
    \includegraphics[width=\linewidth]{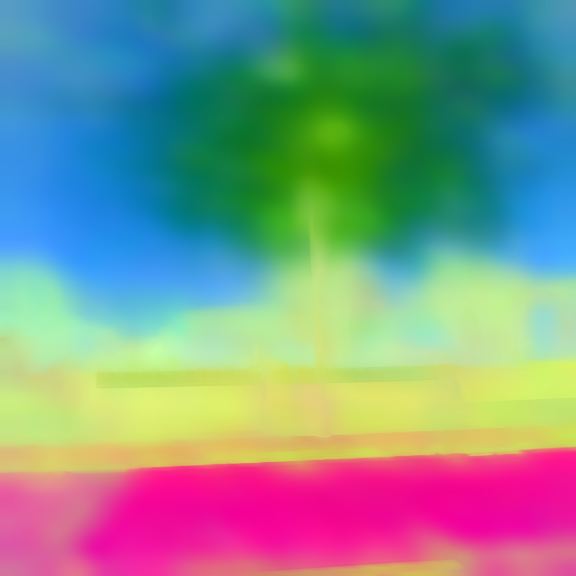}
    \end{subfigure}

    \vspace{1mm}

  \begin{subfigure}{0.32\linewidth}
    \includegraphics[width=\linewidth]{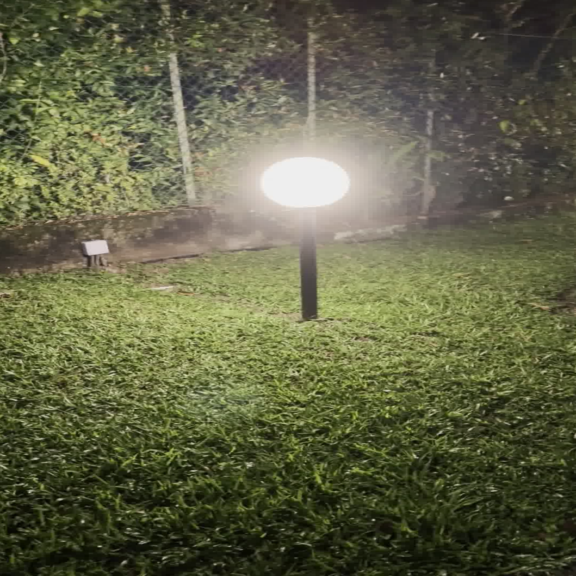}
    \caption{Input}
    \label{fig:short-a}
  \end{subfigure}
  \begin{subfigure}{0.32\linewidth}
    \includegraphics[width=\linewidth]{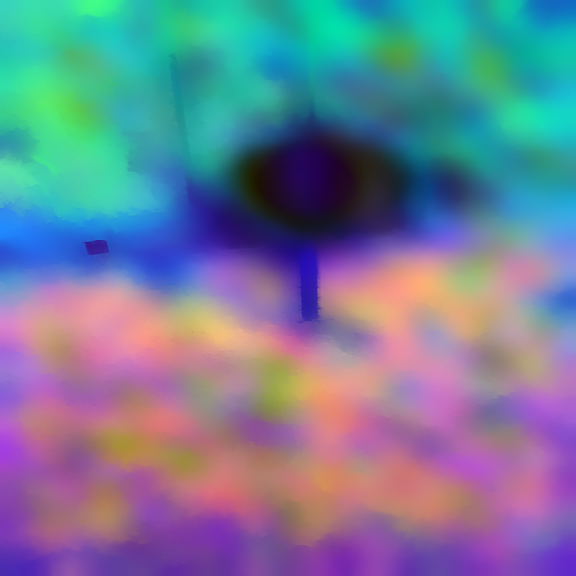}
    \caption{CLIP \cite{radford2021learning}}
    \label{fig:short-b}
  \end{subfigure}
  \begin{subfigure}{0.32\linewidth}
    \includegraphics[width=\linewidth]{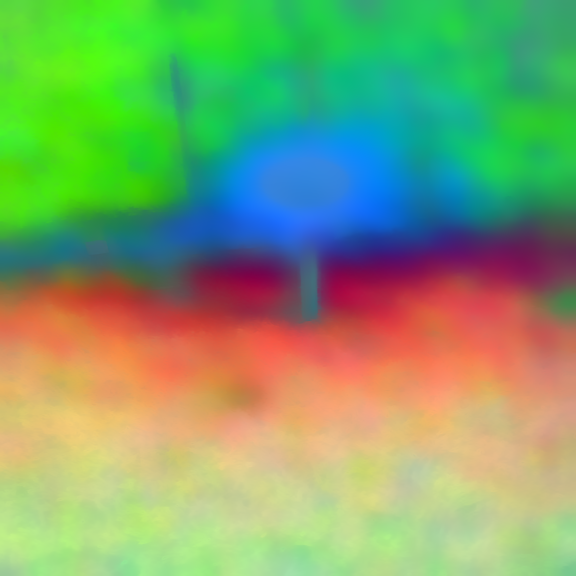}
    \caption{DINO \cite{caron2021emerging}}
    \label{fig:short-a}
  \end{subfigure}

\caption{Visualization of features extracted by different Vision Foundation Models.}

  \label{Fig_Dino}
\end{figure}

Unlike traditional 3DGS, which optimizes only training views, our approach also renders novel views and extracts their semantic features during training.
These features are queried against the semantic feature bank, retrieving the most similar ones by minimizing their distance.
This process allows 3DGS to progressively refine missing structural information through feature optimization.
As shown in Figure~\ref{Fig_trailor}, our method effectively reconstructs nighttime scenes, including glow regions.

As no dataset exists for evaluating 3D scene reconstruction in nighttime settings with strong glow, we introduce \textbf{NightGlow}, a new dataset for this task.
NightGlow consists of 18 scenes with approximately 540 images, all featuring a strong presence of glow.
Both qualitative and quantitative results show that GlowGS effectively reconstructs glow regions in novel views.
Moreover, GlowGS is model-agnostic, making it compatible with various 3DGS techniques, including 3DGS~\cite{kerbl20233d}, CGS~\cite{lee2023compact}, and MGS~\cite{yu2023mip}.
Our key contributions are as follows:
\begin{itemize}  
	\item \textbf{Semantic feature generation:} We introduce a method that combines a diffusion model and Vision Foundation Models (VFM) to generate high-quality semantic features, providing implicit structural cues for novel views.  
	
	\item \textbf{Novel-view semantic learning:} We propose a learning strategy that enables 3D Gaussian Splatting to optimize novel views without ground-truth supervision, enforcing implicit structural constraints to ensure semantically accurate, artifact-free renders.  
	
	\item \textbf{Extensive experiments:} We conduct comprehensive experiments demonstrating that GlowGS significantly improves the reconstruction of real-world nighttime glow scenes. Our method outperforms MGS~\cite{yu2023mip}, achieving a 1.78 improvement in PSNR.  
\end{itemize}

\begin{figure*}
	\centering
	\includegraphics[width=0.99\linewidth]{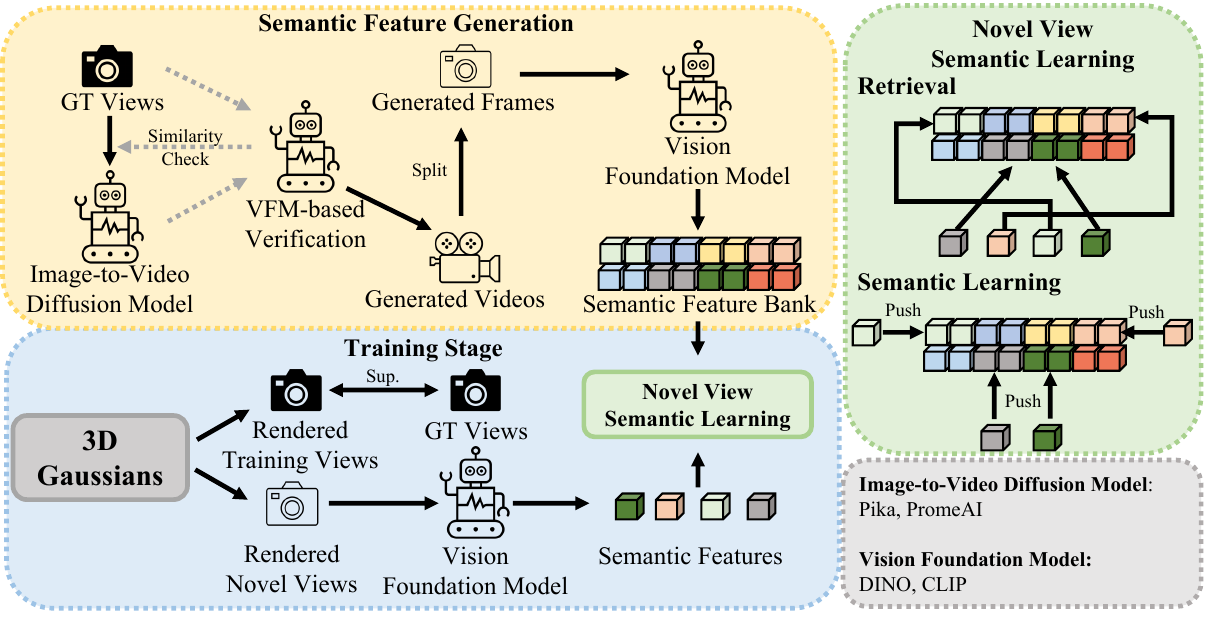}
	\caption{\textbf{Overview of our GlowGS, which centers around two core ideas:} semantic feature generation and novel-view semantic learning.
		Given the training views, our semantic feature generation uses image-to-video diffusion models to create novel views with unknown poses, followed by a VFM-based module to assess their quality. We then extract robust semantic features from the high-quality views and build a semantic feature bank.
		In novel-view semantic learning, we refine these views by extracting their features, retrieving the most similar ones from the bank, and minimizing the distance between them to ensure consistency and quality.
		}
	\label{overview}
\end{figure*}

\section{Related Work}
\label{sec:related_work}
\paragraph{Novel View Synthesis}
Given several captured images with corresponding viewpoints, Novel View Synthesis (NVS) generates new images from different viewpoints~\cite{gortler2023lumigraph,levoy2023light}. 
Neural Radiance Fields (NeRF)\cite{mildenhall2021nerf} achieves this via volume rendering\cite{max2005local,max1995optical}, using MLPs~\cite{mescheder2019occupancy,park2019deepsdf,chen2019learning} to model scenes as continuous functions, providing a compact representation. 
However, its rendering speed is slow due to MLP evaluation at each ray point. To enable real-time rendering, some methods distill a pre-trained NeRF into a sparse representation~\cite{reiser2021kilonerf,hedman2021baking,yu2021plenoctrees,yariv2023bakedsdf}. 
More recently, advanced scene representations have been developed to boost training and rendering efficiency~\cite{fridovich2022plenoxels,kulhanek2023tetra,chen2022tensorf,muller2022instant,xu2022point,chen2023neurbf,kerbl20233d,cao3dot}. Despite these advancements, NeRF-based methods still face slow rendering speeds, limiting their real-time application.

3D Gaussians Splatting (3DGS)~\cite{kerbl20233d, yu2023mip, lee2023compact, Teng2025} has been proposed for 3D scene reconstruction. 
Unlike NeRF, which uses MLPs to model scenes as continuous functions, 3DGS represents scenes with Gaussian distributions, enabling efficient rendering by projecting these Gaussians onto 2D views. 
Lee et al.~\cite{lee2023compact} introduce Compact 3DGS, using a learnable mask strategy to reduce Gaussians while preserving performance. Yu et al.~\cite{yu2023mip} add a 3D smoothing filter to address artifacts from varying sampling rates and replace the 2D dilation filter with a 2D Mip filter to reduce aliasing and dilation.

\vspace{-3mm}
\paragraph{Diffusion Models and Vision Foundation Models} 
Existing large-scale pretrained diffusion models \cite{rombach2022high, peebles2023scalable, lingeocomplete, ho2022video, wang2020imaginator, chen2023control, wu2023tune, pika2024, promeai2024, linrgb, lin2025seeing, lin2024nightrain} and Vision Foundation Models \cite{caron2021emerging, radford2021learning, kirillov2023segment, he2022masked} excel in image and video generation, as well as image understanding.
Specifically, \cite{rombach2022high} introduces a latent-based diffusion model for image generation, while large-scale pretrained models by Pika \cite{pika2024} and PromeAI \cite{promeai2024} have shown remarkable video generation capabilities, extending a single video frame into a high-fidelity 4-second clip.
Vision foundation models like DINO \cite{caron2021emerging} and CLIP \cite{radford2021learning} demonstrate advanced image understanding. Trained using self-supervised strategies, they learn from vast amounts of unlabeled data, leading to enhanced image comprehension.

\section{Proposed Method: GlowGS}
Figure~\ref{overview} provides an overview of GlowGS, highlighting its two key components: semantic feature generation and novel-view semantic learning.
The details are as follows.

\subsection{Semantic Feature Generation} 
\label{sec_mac}
Our semantic feature generation, illustrated in the yellow panel in Figure~\ref{overview}, synthesizes novel views with unknown camera poses using an image-to-video diffusion model. It then extracts their semantic features with a Vision Foundation Model (VFM) and stores them in a semantic feature bank.  
This whole process consists of three core components: image-to-video generation, VFM-based verification, and feature extraction.

\vspace{-2mm}
\paragraph{Image-to-Video Generation} 
Here, we aim to synthesize novel views with unknown camera poses from the training views of a scene using an image-to-video diffusion model~\cite{pika2024, promeai2024}. Given a scene $\mathbf{D} = \{ f^{0}_{i} | i=1,2,..., N \}$, where $f^{0}_{i}$ is the $i$-th training view and $N$ is the number of training views, the generation is formulated as:
\begin{equation} 
	\label{eqn_fga} 
	\mathbf{V}_{i} = {\bf{F}}_{\bf{I2V}} (f^{0}_{i}), 
\end{equation}
where $\bf{F}_{\bf{I2V}}(\cdot)$ is the image-to-video diffusion model, and $\mathbf{V}_{i} = \{ f^{j}_{i} | j=1,2,..., M \}$ represents the set of $M$ generated novel views.

Since the camera poses for each novel view $f^{j}_{i}$ are unknown due to the synthetic and random generation process, they cannot directly guide the training of our 3DGS. The generated views may also suffer from hallucinations, inconsistencies, and low quality.
To address these issues, we introduce a VFM-based verification method to assess the quality of the generated views $\mathbf{V}_{i}$.

\vspace{-2mm}
\paragraph{VFM-based Verification}  
We leverage a VFM, such as DINO~\cite{caron2021emerging}, to evaluate the quality of the generated videos. Our main idea is that the generated views should closely match the semantic content of the original frame. With its image understanding capabilities, the VFM can extract robust semantic features, making it well-suited for assessing the quality and consistency of the generated novel views.

Specifically, we extract the semantic features of the $i$-th training view and its corresponding novel views $\mathbf{V}_{i} = \{ {f}^{j}_{i} \,|\, j = 1,2,\dots,M \}$.
We then compute the $L_{2}$ distance between the features of the training view ${f^{0}_{i}}$ and those of each novel view ${f^{j}_{i}}$:
\begin{equation} 
	\label{Eqn_VFM} 
	D_{i} = \frac{1}{M}\sum^{M}_{n=1}{|| \mathbf{F}_{\mathbf{DINO}}(f^{0}_{i}) - \mathbf{F}_{\mathbf{DINO}}(f^{n}_{i}) ||}_2,
\end{equation}
where $D_{i}$ is the distance between the features of the generated results and of the original image. $\bf{F}_{\bf{DINO}} (\cdot)$ represents DINO’s feature extraction backbone, providing a global feature representation.
If $D_{i}$ exceeds the threshold, the novel view is too different, and we regenerate it until $D_{i}$ falls below the threshold.

By applying this verification process, we expect to obtain high-quality views. Even though minor hallucinations may still remain, our GlowGS can mitigate these effects by using our novel-view semantic learning.
%

\vspace{-2mm}
\paragraph{Feature Extraction} 
Using high-quality views, a VFM extracts features to construct a semantic feature bank. In addition to DINO~\cite{caron2021emerging}, other VFMs, such as CLIP~\cite{radford2021learning}, can be used to extract semantic features. Given a view ${f^{j}_{i}}$, the semantic features are defined as:  
\begin{equation}
	\label{eqn_fga}
	\mathbf{B}_{i,j} = {\bf{F}}_{\bf{VFM}} ({f^{j}_{i}}),
\end{equation}
where ${\bf{F}}_{\bf{VFM}}(\cdot)$ is the feature extraction backbone of the VFM. The output, $\mathbf{B}_{i,j} =  \{ b^{p}_{i,j} | p=1,2,..., P\}$, represents patch-level features, where $P$ is the number of patches, and $b^{p}_{i,j}$ denotes the feature of the $p$-th patch.  
The final semantic feature bank is formulated as $\mathbf{B} =  \{ b^{s} | s=1,2,..., S\}$, where $S$ is the total number of patch-level features.

\begin{table*}[t]
  \centering
  \renewcommand\arraystretch{1.2} 
  \setlength{\tabcolsep}{2.5mm}{
    \begin{tabular}{l|c|c|c|c|c|c|c|c|c}
    \toprule
    \multicolumn{1}{c|}{\multirow{2}[4]{*}{Method}} & \multicolumn{3}{c|}{NightGlow} & \multicolumn{3}{c|}{RawNeRF-Glow (sRGB)} & \multicolumn{3}{c}{Bilarf-Glow} \\
\cmidrule{2-10}          & PSNR $\uparrow$  & SSIM $\uparrow$ & LPIPS $\downarrow$ & PSNR $\uparrow$ & SSIM $\uparrow$ & LPIPS $\downarrow$ & PSNR $\uparrow$ & SSIM $\uparrow$ & LPIPS $\downarrow$ \\
    \midrule
    LLNeRF & 25.32  & 0.7845  & 0.3029  &       20.75 &	0.5847 &	0.4388        &  18.63 &	0.6263 &	0.4033   \\
    \midrule
    AlethNeRF & 24.94  & 0.7767  & 0.3384  &       23.38 	& 0.6424 &	0.4250   &       18.19 &	0.5749 	& 0.5025   \\
    \midrule
    \midrule
    3DGS  & 26.11  & 0.8156  & 0.2378  & 23.04  & 0.6455  & 0.3615  & 17.61  & 0.6171  & 0.3399  \\
    \midrule
    3DGS+\textbf{Ours} & 27.80  & 0.8679  & 0.1869  & 24.39  & 0.6731  & 0.3464  &   18.34 &	0.6921 & 0.2799  \\
    \midrule
    \midrule
    CGS   & 26.20  & 0.8083  & 0.2511  & 23.54  & 0.6319  & 0.3712  & 18.12  & 0.6147  & 0.3295  \\
    \midrule
    CGS+\textbf{Ours} & 27.76  & 0.8627  & 0.1902  & 25.29  & 0.6923  & 0.3482  & 18.99  & 0.6823  & 0.2766  \\
    \midrule
    \midrule
    MGS & 26.46  & 0.8233  & 0.2272  & 23.88  & 0.6618  & 0.3492  & 17.76  & 0.6412  & 0.3051  \\
    \midrule
    MGS+\textbf{Ours} & \textbf{28.24}  & \textbf{0.8739}  & \textbf{0.1847}  & \textbf{25.37}  & \textbf{0.7040}  & \textbf{0.3430}  & \textbf{19.14}  & \textbf{0.7308}  & \textbf{0.2404}  \\
    \bottomrule
    \end{tabular}%
    }
  \caption{Comparison of performance metrics (PSNR, SSIM, and LPIPS) for methods trained on our NightGlow, RawNeRF-Glow \cite{mildenhall2022nerf}, and Bilarf-Glow \cite{wang2024bilateral} datasets. LLNeRF \cite{wang2023lighting} and AlethNeRF \cite{cui2024aleth} are NeRF-based methods, while 3DGS \cite{kerbl20233d}, CGS \cite{lee2023compact}, and MGS \cite{yu2023mip} are 3DGS-based methods. We integrate our GlowGS into three different 3DGS-based backbones for a comprehensive evaluation.}
  \label{tab_quantitative}%
\end{table*}%

\subsection{Novel-View Semantic Learning}
During training, 3DGS models can be optimized if novel views of a scene are available with known camera poses. However, novel views generated by our image-to-video diffusion models lack camera pose information.  
To address this, we propose novel-view semantic learning, illustrated in the green panel in Figure~\ref{overview}. This process enables 3DGS to obtain novel view information through the semantic feature bank.  
The key idea is that similar regions share closely related semantic features despite minor pixel-level differences. Thus, the semantic features in the bank constrain corresponding regions in novel views rendered by 3DGS.  
Note that our novel-view semantic learning is model-agnostic, allowing for the use of different 3DGS models.

In 3DGS, a scene is represented by a set of 3D Gaussians $\{ \mathbf{G}_{k}(x) | k=1,2,..., K\} $, where $K$ is the number of 3D Gaussians. $\mathbf{G}_{k}(x)$ is the $k$-th 3D Gaussian that can be formulated as:
\begin{equation}
\mathbf{G}_{k}(x) = e^{-\frac{1}{2} (x-\mu_{k})^T \Sigma^{-1}_{k} (x-\mu_{k})},
\label{gs}
\end{equation}
where $\mu_k \in \mathbb{R}^{3 \times 1}$ and $\Sigma^{-1}_{k} \in \mathbb{R}^{3 \times 3}$ are the centre and covariance matrix of the $k$-th 3D Gaussian. 
Since the covariance matrices should be positive semi-definite, previous methods \cite{kerbl20233d} formulate $\Sigma_{k}$ as follows:
\begin{equation}
\Sigma_k = R_{k}S_{k}S^{T}_{k}R^{T}_{k},
\label{sigma}
\end{equation}
where $S_{k}$ and $R_{k}$ are the scaling matrix and rotation matrix, respectively.

The color $c(x)$ of a pixel $x$ is built based on the spherical harmonics. It can be rendered by alpha blending, which can be expressed as: 
\begin{equation}
c(x) = \sum_{i \in N} c_i  \alpha_i \mathbf{G}^{\bf{2D}}_{i}(x) \prod_{j=1}^{i-1} (1 -  \alpha_i \mathbf{G}^{\bf{2D}}_{i}(x)),
\label{gs_rendering}
\end{equation}
where $N$ is the number of sorted 2D Gaussians associated with this pixel. $\mathbf{G}^{\bf{2D}}_{i}$ is a 2D Gaussian transformed from the 3D Gaussian $\mathbf{G}_{i}(x)$. The transformation is based on a rotation matrix $\mathbf{R} \in \mathbb{R}^{3 \times 3}$ and a translation matrix $\mathbf{t} \in \mathbb{R}^{3 \times 1}$. $\alpha_i$ is the opacity of the 3D Gaussian $\mathbf{G}_{i}(x)$. 

Given the rotation and translation matrix, the novel view $\mathbf{\overline{V}}$ can be rendered by Eq. \ref{gs_rendering}. Note that,  we do not have the ground truth of novel views. 
We then use a VFM ${\bf{F}}_{\bf{VFM}}(\cdot)$ to extract patch-based semantic features of the novel view:
\begin{equation}
	\label{eqn_fga}
	\mathbf{\overline{B}} = {\bf{F}}_{\bf{VFM}} (\mathbf{\overline{V}}),
\end{equation}
where $\mathbf{\overline{B}} =  \{ \overline{b}^{p} | p=1,2,..., P\} $ is the output of a VFM. $\overline{b}^{p}$ represents the semantic features of the $p$-th patch. 

After obtaining the semantic features of the novel view,  we utilize our semantic feature bank to guide their optimization. The process consists of two steps: retrieval and semantic learning.  
Our retrieval process aims to search for the most similar semantic features $b^{m  \mid p}$ in the bank $\mathbf{B}$ for each feature $\overline{b}^{p}$. 
Once the most similar semantic features are found, we have paired semantic features. Our semantic learning process forces our model to reduce the distance between the paired semantic features:
\begin{equation}
	\label{eqn_fga}
	\mathcal{L} = \lambda \mathcal{L}_{\bf{ori}} + (1-\lambda)\frac{1}{P}\sum^{P}_{p=1}{|| \overline{b}^{p} - b^{m \mid p} ||}_2,
\end{equation}
where $\lambda$ is the weight that balances different losses. $\mathcal{L}_{\bf{ori}}$ is the original loss of our backbones. 
Since novel-view semantic learning loss does not modify 3DGS backbones, GlowGS can be applied to various 3DGS models~\cite{kerbl20233d, yu2023mip, lee2023compact}. 

\section{Experiments}
Since no dataset is designed for evaluating 3D scene reconstruction in nighttime glow scenes, particularly with strong glow, we introduce the NightGlow dataset. We also identify nighttime scenes with light glow effects in existing datasets, such as RawNeRF~\cite{mildenhall2022nerf} and Bilarf~\cite{wang2024bilateral}, and include them for evaluation.

Our NightGlow dataset consists of 18 scenes, each with approximately 30 images. All scenes feature night glow effects, with most exhibiting strong glow. Following the settings from~\cite{kerbl20233d}, we construct the dataset and will publicly release it.  
From RawNeRF-Glow~\cite{mildenhall2022nerf}, we select six scenes: candlefiat, gardenlights, notchbush, parkstatue, stove, and streetcorner. These scenes contain light glow effects. RawNeRF is originally designed for 3D reconstruction in the raw space, but since our method focuses on well-lit night images, particularly glow regions, we use sRGB images for our experiments.  
From Bilarf-Glow~\cite{wang2024bilateral}, we select three nighttime scenes: building, pondbike, and strat. Each contains light glow effects.  
To evaluate our method, we benchmark against three state-of-the-art (SOTA) 3D Gaussian Splatting methods~\cite{kerbl20233d, yu2023mip, lee2023compact} and conduct experiments on these datasets.

\begin{figure*}[t]
\centering

\includegraphics[width=0.18\textwidth, height=0.25\textwidth]{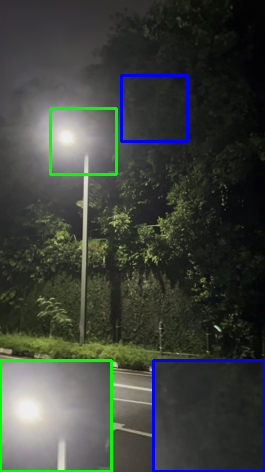}
\hspace{1mm}
\includegraphics[width=0.18\textwidth, height=0.25\textwidth]{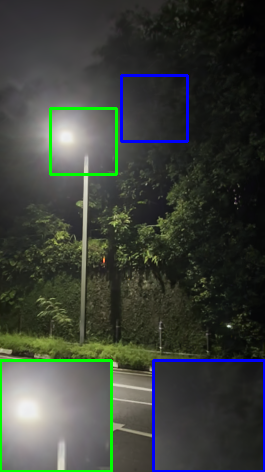}
\hspace{1mm}
\includegraphics[width=0.18\textwidth, height=0.25\textwidth]{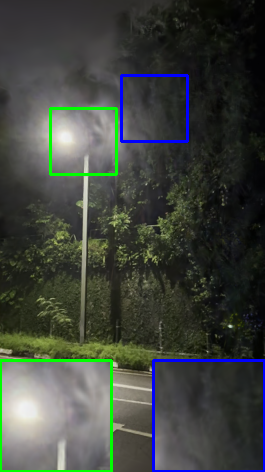}
\hspace{1mm}
\includegraphics[width=0.18\textwidth, height=0.25\textwidth]{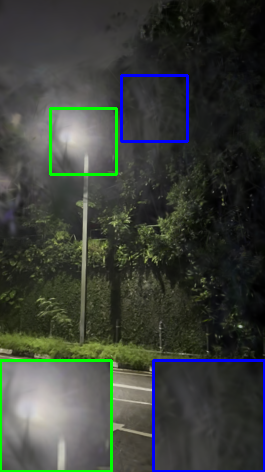}
\hspace{1mm}
\includegraphics[width=0.18\textwidth, height=0.25\textwidth]{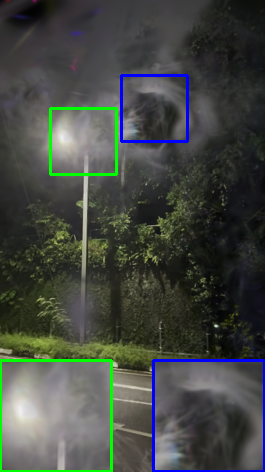}



\vspace{1mm}

\includegraphics[width=0.18\textwidth, height=0.25\textwidth]{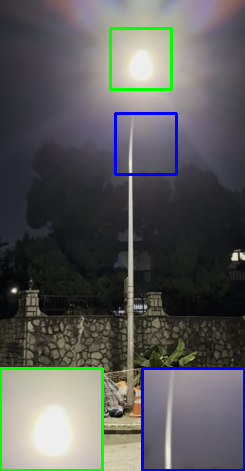}
\hspace{1mm}
\includegraphics[width=0.18\textwidth, height=0.25\textwidth]{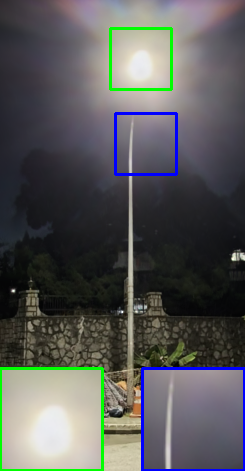}
\hspace{1mm}
\includegraphics[width=0.18\textwidth, height=0.25\textwidth]{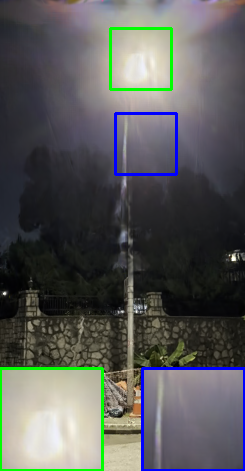}
\hspace{1mm}
\includegraphics[width=0.18\textwidth, height=0.25\textwidth]{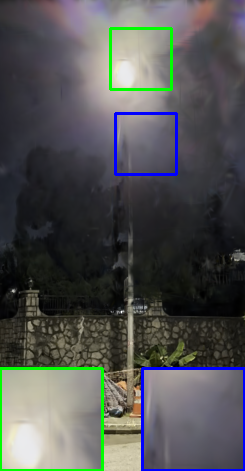}
\hspace{1mm}
\includegraphics[width=0.18\textwidth, height=0.25\textwidth]{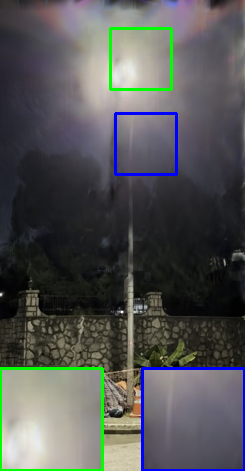}

\vspace{1mm}
\begin{subfigure}[t]{0.18\textwidth}
    \includegraphics[width=1\textwidth, height=1\textwidth]{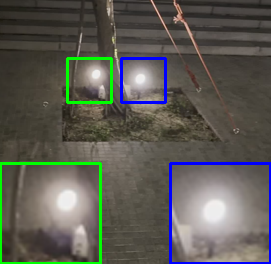}
    \caption{GT}
\end{subfigure}
\hspace{1mm}
\begin{subfigure}[t]{0.18\textwidth}
    \includegraphics[width=1\linewidth, height=1\textwidth]{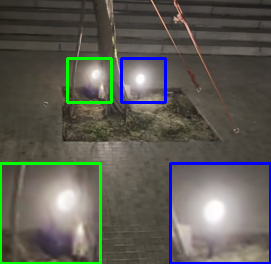}
    \caption{Ours}
\end{subfigure}
\hspace{1mm}
\begin{subfigure}[t]{0.18\textwidth}
    \includegraphics[width=1\linewidth, height=1\textwidth]{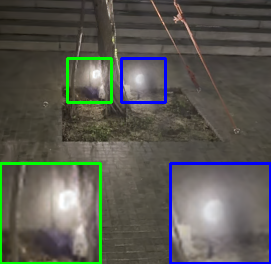}
    \caption{3DGS}
\end{subfigure}
\hspace{1mm}
\begin{subfigure}[t]{0.18\textwidth}
    \includegraphics[width=1\linewidth, height=1\textwidth]{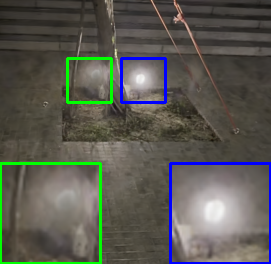}
    \caption{CGS}
\end{subfigure}
\hspace{1mm}
\begin{subfigure}[t]{0.18\textwidth}
    \includegraphics[width=1\linewidth, height=1\textwidth]{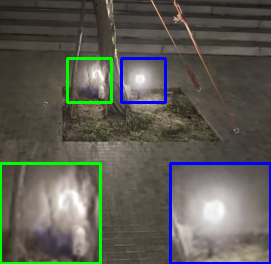}
    \caption{MGS}
\end{subfigure}

\caption{Qualitative results from 3DGS \cite{kerbl20233d}, CGS \cite{lee2023compact}, MGS \cite{yu2023mip} and our method, on nighttime glow scenes. All results are from novel views. Our method not only preserves the details of night images but also effectively reconstructs glow regions. Zoom in for better visualization.}
\label{Fig_exp1}
\end{figure*}

\subsection{Implementation Details}
\label{sec_id}
For each scene, we use six views for training and the rest for evaluation. To ensure fair comparisons, we retain the original optimal settings of baseline methods~\cite{kerbl20233d, yu2023mip, lee2023compact}, including the optimizer, learning rate, and other hyperparameters. We then integrate GlowGS into these backbones to enhance novel view learning.  
All experiments run on an A5000 GPU with 24 GB of RAM. To balance the original loss with our novel-view learning loss, we introduce a weight parameter $\lambda$, fixed at 0.01 in all experiments.  
GlowGS is driven by two core ideas: semantic feature generation and novel-view semantic learning.

\vspace{-2mm}
\paragraph{\textbf{Semantic Feature Generation}} 
In semantic feature generation, we use image-to-video diffusion models to synthesize novel views and a VFM to assess their quality. Once high-quality views are obtained, we extract their semantic features using VFMs. For our experiments, we use Pika~\cite{pika2024} and PromeAI~\cite{promeai2024} as image-to-video diffusion models. Pika generates 3-second videos per input view, while PromeAI generates 4-second videos. We extract one frame per second, yielding three frames from Pika and four from PromeAI for each training view.  

For VFM-based verification, we use DINO to measure the distance between input and generated views. If the distance exceeds 1.5, the image-to-video diffusion models re-generate novel views by adjusting the motion intensity or random seed.  
Once high-quality novel views are obtained, we extract robust semantic features using an VFM. We try a few models such as   DINO~\cite{caron2021emerging}, CLIP~\cite{radford2021learning}, and ViT~\cite{dosovitskiy2020image, fu2024featup}.  

\vspace{-2mm}
\paragraph{\textbf{Novel-View Semantic Learning}} 
During training, we generate new camera poses by perturbing the rotation and translation matrices of the training views. After splatting, novel views are obtained. Since some VFMs support only square inputs, we randomly crop a $256 \times 256$ region from each novel view before feeding it into VFMs for semantic feature extraction. The number of patches, $P$, depends on the output resolution of the selected VFM. 

\begin{figure*}[t]
\centering

\includegraphics[width=0.18\textwidth, height=0.25\textwidth]{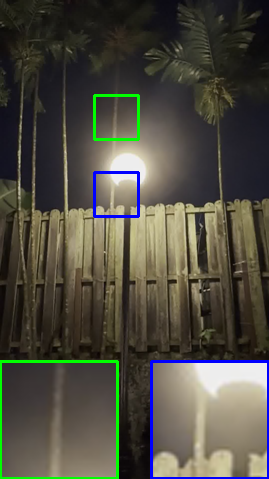}
\hspace{1mm}
\includegraphics[width=0.18\textwidth, height=0.25\textwidth]{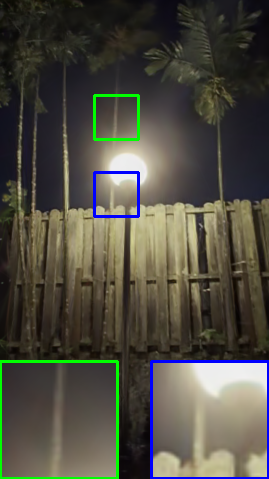}
\hspace{1mm}
\includegraphics[width=0.18\textwidth, height=0.25\textwidth]{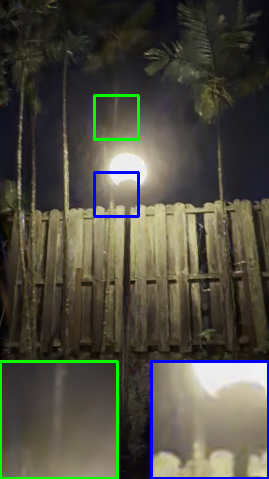}
\hspace{1mm}
\includegraphics[width=0.18\textwidth, height=0.25\textwidth]{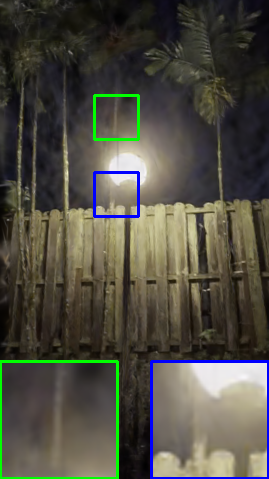}
\hspace{1mm}
\includegraphics[width=0.18\textwidth, height=0.25\textwidth]{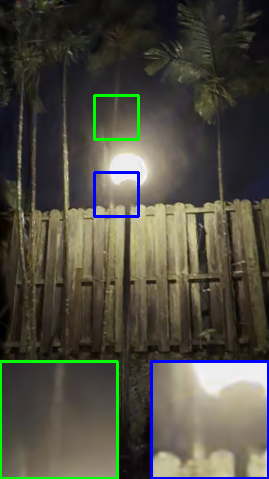}

\vspace{1mm}

\begin{subfigure}[t]{0.18\textwidth}
    \includegraphics[width=1\textwidth, height=1.4\textwidth]{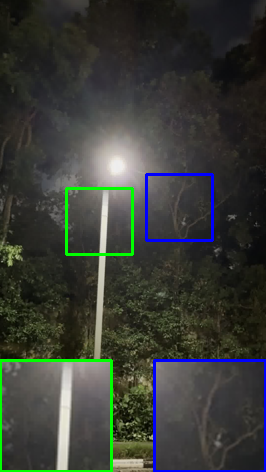}
    \caption{GT}
\end{subfigure}
\hspace{1mm}
\begin{subfigure}[t]{0.18\textwidth}
    \includegraphics[width=1\linewidth, height=1.4\textwidth]{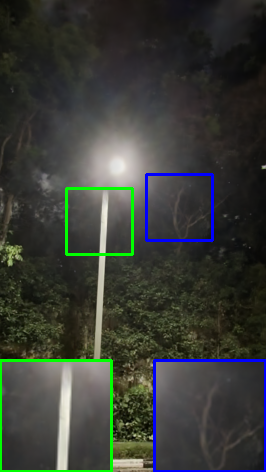}
    \caption{Ours}
\end{subfigure}
\hspace{1mm}
\begin{subfigure}[t]{0.18\textwidth}
    \includegraphics[width=1\linewidth, height=1.4\textwidth]{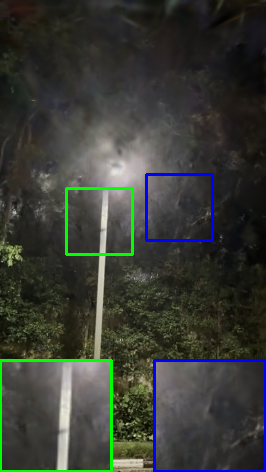}
    \caption{3DGS}
\end{subfigure}
\hspace{1mm}
\begin{subfigure}[t]{0.18\textwidth}
    \includegraphics[width=1\linewidth, height=1.4\textwidth]{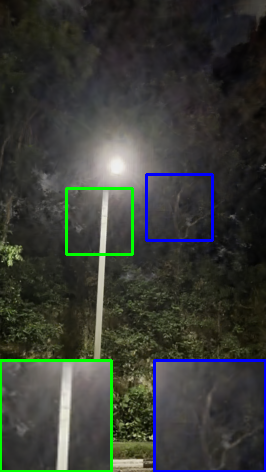}
    \caption{CGS}
\end{subfigure}
\hspace{1mm}
\begin{subfigure}[t]{0.18\textwidth}
    \includegraphics[width=1\linewidth, height=1.4\textwidth]{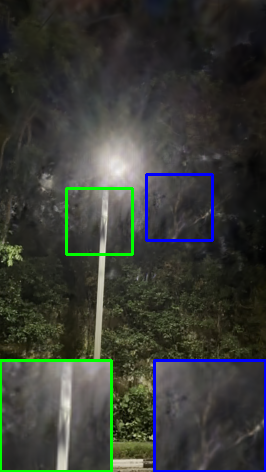}
    \caption{MGS}
\end{subfigure}

\caption{Qualitative results from 3DGS \cite{kerbl20233d}, CGS \cite{lee2023compact}, MGS \cite{yu2023mip} and our method, on nighttime glow scenes. All results are from novel views. Our method not only preserves the details of night images but also effectively reconstructs glow regions. Zoom in for better visualization.}
\label{Fig_exp2}
\end{figure*}

\subsection{Quantitative Evaluation} 
The quantitative results in Table~\ref{tab_quantitative} compare performance across three datasets: NightGlow, RawNeRF-Glow~\cite{mildenhall2022nerf}, and Bilarf-Glow~\cite{wang2024bilateral}. While RawNeRF-Glow and Bilarf-Glow contain only light glow scenes, NightGlow includes strong glow effects. Integrating existing methods with GlowGS consistently improves performance across all datasets.

For the NightGlow dataset, we benchmark existing 3D scene reconstruction methods, including both NeRF-based and 3DGS-based approaches. For NeRF-based methods, we select LLNeRF~\cite{wang2023lighting} and AlethNeRF~\cite{cui2024aleth}, as they are designed for low-light and nighttime conditions.
We also apply GlowGS to several 3D Gaussian Splatting methods, including 3DGS~\cite{kerbl20233d}, CGS~\cite{lee2023compact}, and MGS~\cite{yu2023mip}, and compare their performance with and without novel-view semantic learning. The results, shown in Table~\ref{tab_quantitative}, indicate that MGS achieves an average PSNR of 26.46, surpassing 3DGS and CGS by 0.35 and 0.26, respectively.

Methods with GlowGS achieve significant performance gains. MGS with GlowGS attains an average PSNR of 28.24, surpassing the original MGS by 1.78. Similarly, GlowGS improves 3DGS and CGS by 1.56 and 1.69, respectively.
This improvement comes from GlowGS using image-to-video diffusion models and a VFM-based verification module to generate high-quality novel views. These views are processed by VFMs to build a semantic feature bank, enabling 3DGS models to optimize novel view semantics and boost performance.
GlowGS consistently outperforms baseline 3DGS methods. The VFM-based verification module ensures high-quality generated views, while novel-view semantic learning enhances model training. These components work together to enable effective learning from novel views, driving superior results.

\subsection{Qualitative Evaluation} 

Figure~\ref{Fig_exp1} and Figure~\ref{Fig_exp2} present the qualitative results of 3DGS, CGS, MGS, and our method. As shown, 3DGS, CGS, and MGS often struggle with accuracy, particularly in glow regions, where they produce floating artifacts and degrade performance.
This issue arises because glow regions lack distinct structural features. While 3D Gaussians fit these regions well in training views, they fail to generalize to novel views, leading to artifacts in novel view rendering.

GlowGS uses semantic feature generation to produce high-quality novel views, rich in semantic information despite unknown camera poses. We extract robust features from these views using vision foundation models.
While image-to-video diffusion models introduce some randomness, we use a VFM-based verification module to filter out major inaccuracies and hallucinations. Minor hallucinations are acceptable, as our novel-view semantic learning enables 3DGS to selectively use valuable features for optimization.
These two components significantly enhance 3DGS methods, with GlowGS successfully reconstructing glow regions, as shown in Figure~\ref{Fig_exp1} and Figure~\ref{Fig_exp2}.

\begin{table*}[t]
  \centering
  \setlength{\tabcolsep}{5.2mm}{
    \begin{tabular}{cc|ccc|c|c|c}
    \toprule
    \multicolumn{2}{c}{\textbf{I2V Diffusion Models}} & \multicolumn{3}{c}{\textbf{Vision Foundation Models}} & \multicolumn{3}{c}{\textbf{Metrics}} \\
    \multicolumn{1}{l}{\textbf{I2V-PIKA}} & \multicolumn{1}{l}{\textbf{I2V-PAI}} & \multicolumn{1}{l}{\textbf{ViT}} & \multicolumn{1}{l}{\textbf{CLIP}} & \multicolumn{1}{l}{\textbf{DINO}} & \textbf{PSNR $\uparrow$} & \textbf{SSIM$\uparrow$} & \textbf{LPIPS $\downarrow$} \\
    \midrule
    &       &       &       &       & 26.46  & 0.8233  & 0.2272  \\
    \midrule
    \midrule
    \checkmark     &       & \checkmark      &       &       & 27.47 &	0.8657 &	0.1941   \\
    \checkmark      &       &       & \checkmark      &       & 27.63  & 0.8696  & 0.2015  \\
    \checkmark      &       &       &       & \checkmark      & 28.00  & 0.8724 & \textbf{0.1844}  \\
    & \checkmark      & \checkmark      &       &       & 27.34  &	0.8605 & 	0.1994   \\
          & \checkmark      &       & \checkmark      &       & 27.54  & 0.8644  & 0.2042  \\
          & \checkmark      &       &       & \checkmark      &   \textbf{28.24} &	\textbf{0.8739} &	0.1847  \\
    \end{tabular}%
    }
  \caption{Comparison of performance metrics (PSNR, SSIM, and LPIPS) for methods trained on our NightGlow dataset. `I2V' denotes the image-to-video diffusion models. We use ViT \cite{dosovitskiy2020image, fu2024featup}, CLIP \cite{radford2021learning}, and DINO \cite{caron2021emerging} as vision foundation models. The first row shows the performance of the original MGS \cite{yu2023mip}. 
  }
  \label{tab_abl}%
\end{table*}%

\subsection{Ablation Studies}
\label{sec_abl}
To evaluate the effectiveness of our semantic feature generation and novel-view semantic learning, we conducted ablation studies on the NightGlow dataset.

\vspace{-2mm}
\paragraph{\textbf{Analysis of Different Image-to-Video Diffusion Models}} 
In this paper, we use image-to-video diffusion models, such as Pika~\cite{pika2024} and PromeAI~\cite{promeai2024}, to generate novel views with unknown camera poses. We conducted ablation studies with semantic feature banks generated by different diffusion models. MGS~\cite{yu2023mip} serves as the baseline method, and the results are shown in Table~\ref{tab_abl}.
The results demonstrate that models trained with different semantic feature banks outperform the baseline. For instance, the baseline achieves a PSNR of 26.46, while models using ``I2V-PIKA" and ``I2V-PAI" reach PSNRs of 28.00 and 28.24, respectively, surpassing the baseline by 1.54 dB and 1.78 dB. This improvement is mainly due to our semantic feature bank, which provides rich semantic information from novel views, enabling GlowGS to effectively model nighttime scenes.

We also observe that the baseline achieves similar performance improvements across different image-to-video diffusion models. This is due to the VFM-based verification module, which monitors the quality of generated novel views. If a view's score falls below expectations, it is regenerated, ensuring stable performance improvements across various generative models.
During training, we first retrieve the most similar semantic features and use them to optimize novel views. This approach allows 3DGS to leverage the most effective semantic features for optimization, ensuring consistent performance gains.

\vspace{-2mm}
\paragraph{\textbf{Analysis of Different Vision Foundation Models}} 
In our experiments, we use various vision foundation models, including DINO~\cite{caron2021emerging}, CLIP~\cite{radford2021learning}, and ViT~\cite{dosovitskiy2020image, fu2024featup}, to extract semantic features. To evaluate their effectiveness, we conduct ablation studies on the NightGlow dataset, using MGS~\cite{yu2023mip} as the baseline. The results are presented in Table~\ref{tab_abl}.
The findings show that MGS, combined with different vision foundation models, significantly enhances rendering quality. For example, MGS with DINO achieves a PSNR of 28.24, surpassing the baseline by 1.78 dB. This improvement is due to GlowGS, which enables 3DGS to learn from novel views without requiring camera pose information.

We also observe that "MGS+DINO" achieves the highest average PSNR. For instance, "MGS+DINO" reaches a PSNR of 28.24, 0.9 dB higher than "MGS+ViT." This is largely because DINO, trained on extensive self-supervised data, captures robust feature representations even in night glow scenes. These strong semantic features enhance the novel-view learning process, resulting in superior performance.

\vspace{-2mm}
\paragraph{\textbf{Analysis of Different Baselines}} 
GlowGS is a model-agnostic method, as our novel-view learning process does not require modifying the backbones of 3D Gaussian Splatting methods. To assess its effectiveness, we conduct experiments with various baselines, and the results are shown in Table~\ref{tab_quantitative}.
As expected, integrating GlowGS with 3DGS baselines leads to substantial performance improvements. Specifically, the average PSNRs of 3DGS, CGS, and MGS are 26.11, 26.20, and 26.46, respectively. With GlowGS, these methods improve by 1.69, 1.56, and 1.78 PSNR, respectively.

This demonstrates that GlowGS enhances novel-view learning for existing 3DGS methods in nighttime scenes. As shown in Figure~\ref{Fig_exp1} and Figure~\ref{Fig_exp2}, our method produces high-quality novel views.
The performance gains primarily result from two key factors: semantic feature generation and novel-view semantic learning. Semantic feature generation uses image-to-video diffusion models and vision foundation models to extract features from novel views with unknown camera poses, while novel-view semantic learning optimizes these features. Together, these components drive significant performance improvements.

\section{Conclusion}
We introduced GlowGS, a novel nighttime 3D Gaussian Splatting method that integrates semantic feature generation and novel-view semantic learning.  
Our semantic feature generation synthesizes high-quality features as implicit structural cues for novel-view learning. Specifically, we use image-to-video diffusion models to generate novel views from training views, assessing their quality with a vision foundation model. We then extract semantic features from the high-quality generated views to build a semantic feature bank.  
Our novel-view semantic learning leverages this feature bank to optimize novel views rendered by 3DGS. For each rendered view, we extract its semantic features, search the bank for the most similar features, and minimize their distance. This allows 3DGS to compensate for missing structural cues and ensures semantically coherent, artifact-free renders.  
Experiments on the NightGlow dataset show that integrating GlowGS with existing 3DGS models leads to significant performance improvements.  

{
    \small
    \bibliographystyle{ieeenat_fullname}
    \bibliography{main}
}

\clearpage
\appendix

\section{Visualization}
This paper harnesses vision foundation models to help the reconstruction of nighttime glow scenes. The motivation is that VFMs can generate discriminative representations in glow regions. We show more visualization results in Figure \ref{Fig_Dino}.

\section{Experimental Details}
In this paper, we leverage image-to-video diffusion models and vision foundation models (VFMs) to enable our 3DGS framework to effectively reconstruct nighttime glow scenes. Given a training view, we first use image-to-video diffusion models to generate novel views with unknown camera poses. A VFM-based verification module then assesses the quality of these novel views. Once the high-quality generated views are obtained, we extract robust semantic features using VFMs to construct a semantic feature bank. The experimental details are as follows:

\paragraph{Image-to-Video Diffusion Models} We employ state-of-the-art image-to-video diffusion models, such as Pika \cite{pika2024} and PromeAI \cite{promeai2024}, to synthesize novel views. Pika and PromeAI generate 3-second and 4-second videos per input view, respectively. We then extract one frame per second, resulting in three frames from Pika and four from PromeAI for each training view. For Pika, we do not use a text prompt to guide the generation process, whereas for PromeAI, we use the prompt: ``Slow camera movement, static scene, no new objects." 

\begin{figure*}[t!]
    \centering
    \includegraphics[width=0.9\linewidth]{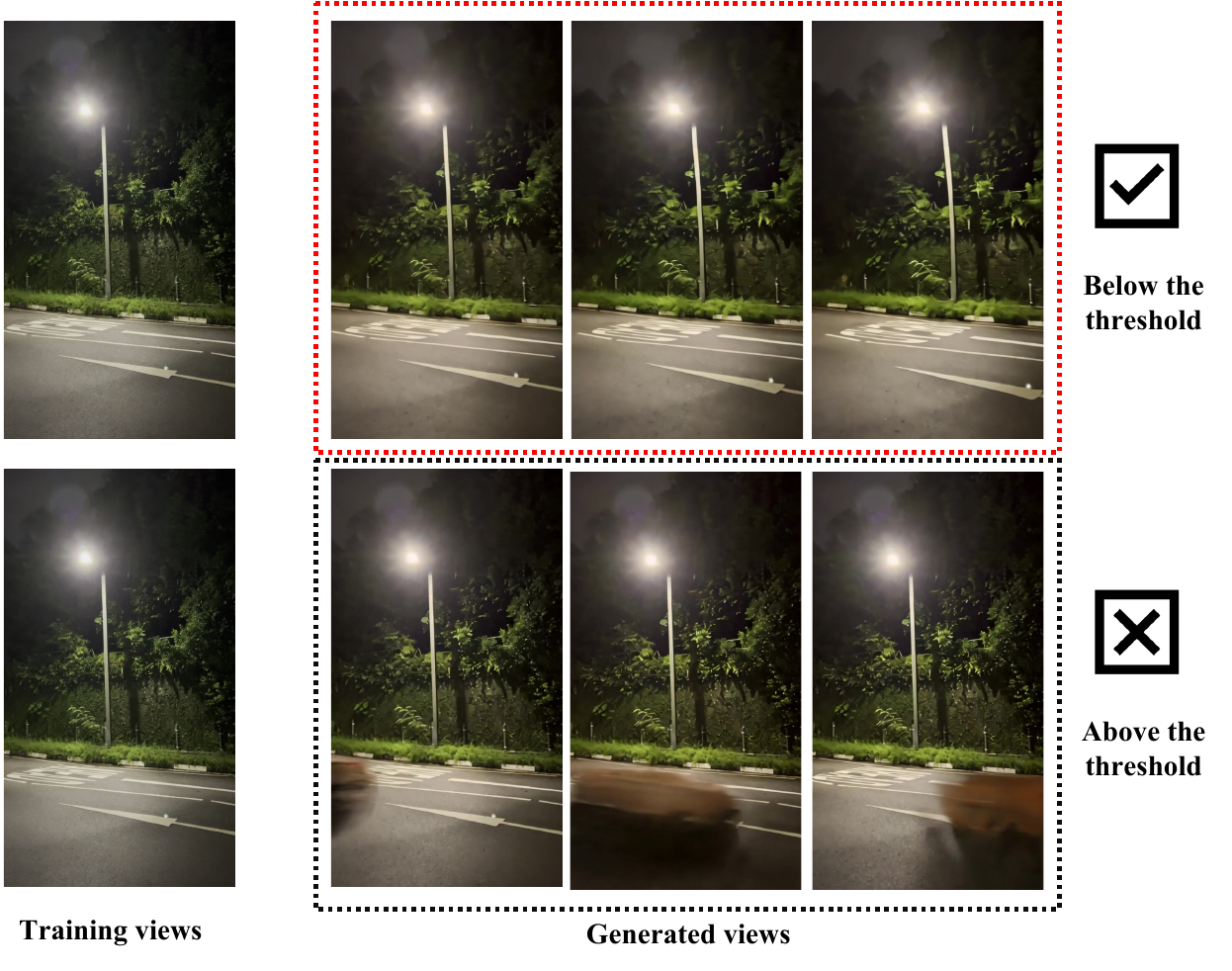}
    \caption{Visualization of generated results. Given a training view, we use image-to-video diffusion models to synthesize novel views. The last three columns show the outputs of these models. Red bounding boxes highlight high-quality results below the threshold, while blue bounding boxes indicate low-quality results exceeding the threshold. 
   }
    \label{fakevideos1}
\end{figure*}

\paragraph{VFM-Based Verification}
We use DINO to measure the distance between the input and generated views, as shown in Figure \ref{fakevideos1}. If the distance exceeds 1.5, the image-to-video diffusion models re-generate novel views by adjusting the motion intensity or random seed. Specifically, for PromeAI, both motion intensity and the random seed can be adjusted, whereas for Pika, only the random seed can be modified.

\paragraph{Feature Extraction}
We leverage vision foundation models such as DINO \cite{caron2021emerging} and CLIP \cite{radford2021learning} to extract semantic features, which are then stored in a semantic feature bank.

\begin{figure}
  \centering
    \begin{subfigure}{0.32\linewidth}
    \includegraphics[width=\linewidth]{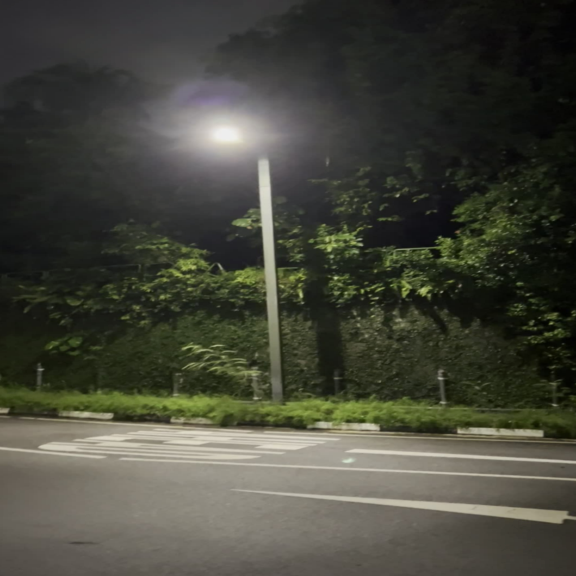}
    \end{subfigure}
    \begin{subfigure}{0.32\linewidth}
    \includegraphics[width=\linewidth]{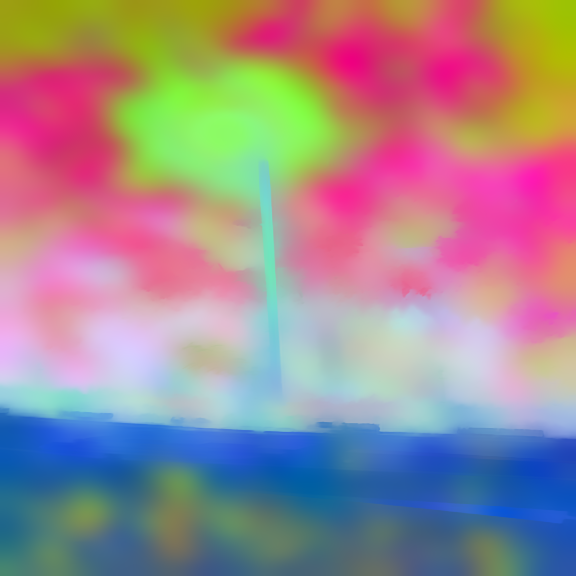}
    \end{subfigure}
    \begin{subfigure}{0.32\linewidth}
    \includegraphics[width=\linewidth]{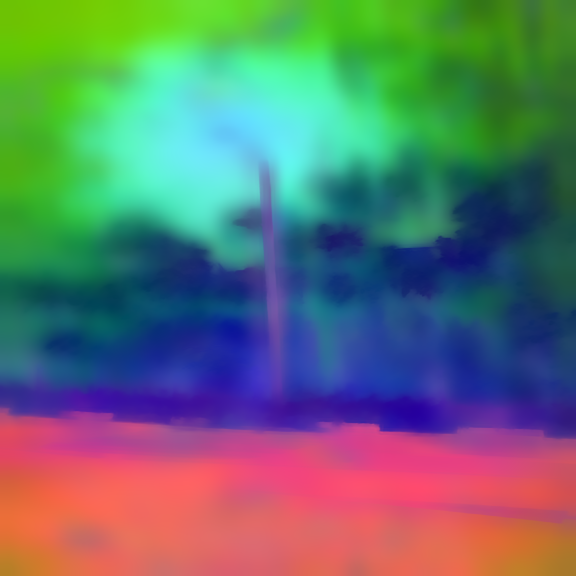}
    \end{subfigure}


    \begin{subfigure}{0.32\linewidth}
    \includegraphics[width=\linewidth]{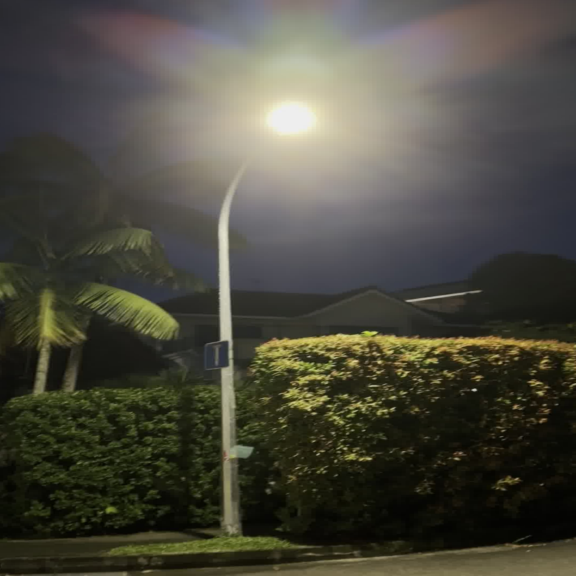}
    \caption{Input}
    \end{subfigure}
    \begin{subfigure}{0.32\linewidth}
    \includegraphics[width=\linewidth]{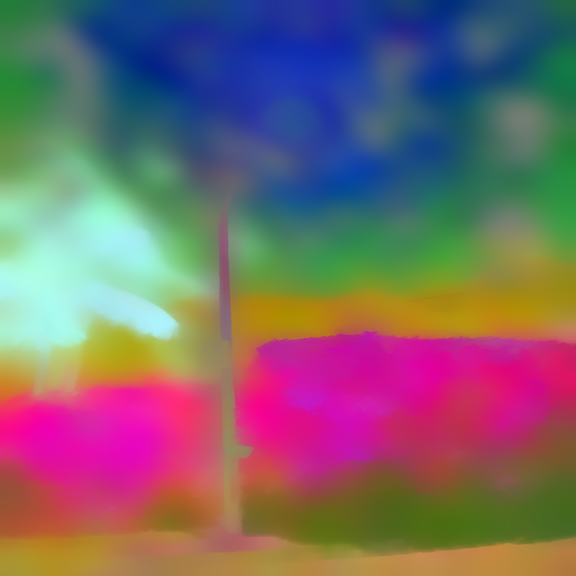}
    \caption{CLIP}
    \end{subfigure}
    \begin{subfigure}{0.32\linewidth}
    \includegraphics[width=\linewidth]{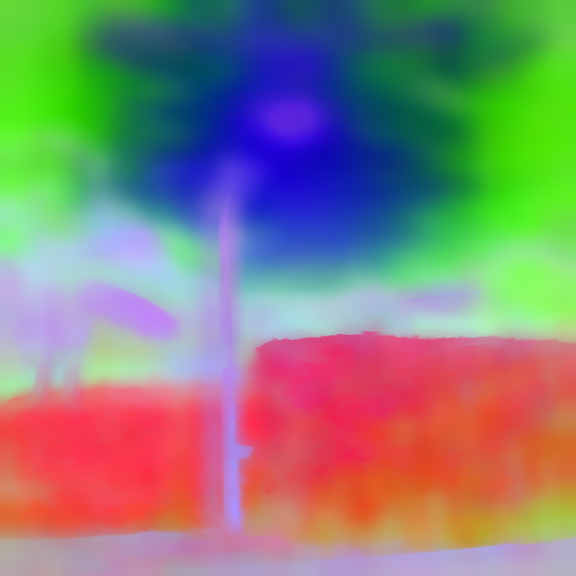}
    \caption{DINO}
    \end{subfigure}


\caption{Visualization of features extracted by different Vision Foundation Models.}

  \label{Fig_Dino}
\end{figure}

\begin{table}[t!]
\centering
\renewcommand{\arraystretch}{1.2}
\caption{(PSNR/SSIM) vs. Number of Training Views}
\label{table_views}
\begin{adjustbox}{max width=\columnwidth}
\begin{tabular}{ccccccc}
\hline
\textbf{Method} & 2 views & 4 views & 6 views & 8 views  & 10 views \\ \hline
\textbf{MGS~\cite{yu2023mip}}     &   21.0/0.64     & 	25.8/0.80  &	26.5/0.82  &	27.6/0.84 	 & 28.2/0.85              \\ \hline
\textbf{MGS~\cite{yu2023mip} + Ours}     &  \textbf{22.1/0.72}      & 	\textbf{27.2/0.85}  &	\textbf{28.2/0.87}  &	\textbf{29.2/0.88} 	 & \textbf{29.8/0.90}             \\ \hline
\end{tabular}
\end{adjustbox}
\end{table}

\begin{figure*}
  \centering
    \begin{subfigure}{0.24\linewidth}
    \includegraphics[width=0.90\linewidth]{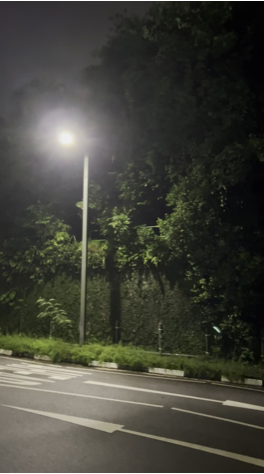}
    \caption{Input}
    \end{subfigure}
    \hfill
    \begin{subfigure}{0.24\linewidth}
    \includegraphics[width=0.90\linewidth]{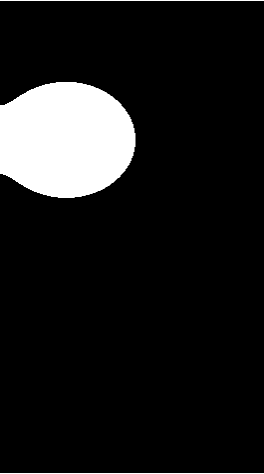}
    \caption{Glow Mask}
    \end{subfigure}
    \hfill
    \begin{subfigure}{0.24\linewidth}
    \includegraphics[width=0.90\linewidth]{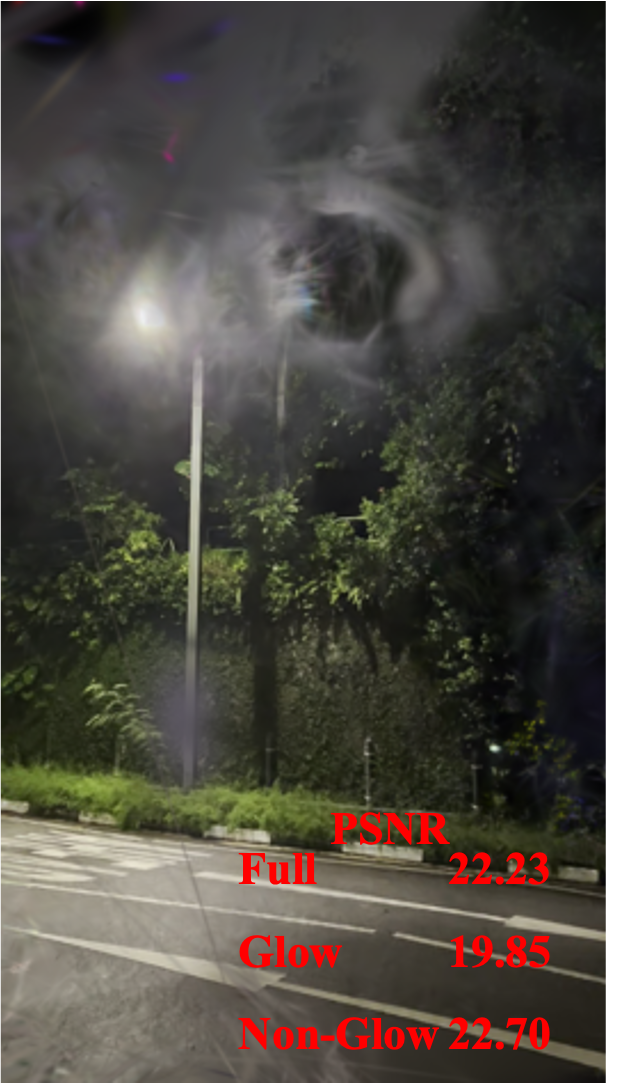}
    \caption{MGS~\cite{yu2023mip}}
    \end{subfigure}
    \hfill
    \begin{subfigure}{0.24\linewidth}
    \includegraphics[width=0.90\linewidth]{Fig_glow/Fig_glow_mgs.png}
    \caption{Ours}
    \end{subfigure}

  \caption{Glow masks and results from MGS~\cite{yu2023mip} and ours.}

  \label{Fig_vis}
\end{figure*}

\begin{table*}[htbp]
    \centering
    \renewcommand{\arraystretch}{1.2}

    \caption{(PSNR $\uparrow$ / SSIM $\uparrow$ / LPIPS $\downarrow$) computed separately on glow and non-glow regions using the glow mask.}
    \begin{adjustbox}{max width=\textwidth}
    \begin{tabular}{c|c|c|c|c|c|c}
    \toprule
    \multirow{2}[4]{*}{\textbf{Method}} & \multicolumn{2}{c|}{NightGlow} & \multicolumn{2}{c|}{RawNerf-Glow (sRGB) ~\cite{mildenhall2022nerf}} & \multicolumn{2}{c}{Bilarf-Glow~\cite{wang2024bilateral}} \\
\cmidrule{2-7}
 & Glow Regions & Non-Glow Regions & Glow Regions & Non-Glow Regions& Glow Regions & Non-Glow Regions \\
    \midrule

\textbf{MGS~\cite{yu2023mip}} 
& 25.7 / 0.97 / 0.2154 
& 26.9 / 0.84 / 0.2298 
& 18.4 / 0.97 / 0.3723 
& 24.7 / 0.67 / 0.3491 
& 17.8 / 0.93 / 0.2853 
& 18.1 / 0.69 / 0.3098 \\
\midrule

\textbf{MGS~\cite{yu2023mip}+ours} 
& \textbf{28.1 / 0.98 / 0.1610}
& \textbf{28.6 / 0.89 / 0.1893}
& \textbf{22.0 / 0.98 / 0.3189}
& \textbf{26.0 / 0.71 / 0.3446}
& \textbf{18.4 / 0.94 / 0.2292}
& \textbf{19.9 / 0.78 / 0.2418} \\
\bottomrule
    \end{tabular}
    \end{adjustbox}
  \label{table_glow}
\end{table*}

\section{Evaluation on Glow and Non-Glow Regions}

Glow is defined as the area surrounding a light source where luminance gradually decays but remains above a certain threshold (e.g., 10\% of the source peak). In practice, this region can be approximated using an Atmospheric Point Spread Function (APSF)~\cite{jin2023enhancing} centered on the detected light source. Fig.~\ref{Fig_vis} illustrates the resulting glow mask. Based on the generated glow mask, PSNR and SSIM are computed separately for the glow and non-glow regions. Table~\ref{table_glow} shows that our method achieves significant improvements in both regions.

\section{Ablation Studies}
In this section, we present additional ablation studies to verify the effectiveness of GlowGS.

\paragraph{Analysis of the Number of Training Views}
In our experimental setup, each scene includes six training views, with the remaining frames used for evaluation. To assess the robustness of GlowGS, we perform ablation studies with varying numbers of training views. As shown in Table \ref{table_views}, our method consistently outperforms baseline approaches, regardless of the number of training views. This improvement stems from our novel-view semantic learning strategy, which optimizes rendered novel views without requiring ground-truth supervision. 

\paragraph{Analysis of Generated Views}
The additional generated views total 18 for Pika and 24 for PromeAI. To assess the impact of the number of generated views, we conduct experiments using 0, 12, and 24 additional views. Our method achieves PSNR/SSIM scores of 26.46/0.8233, 27.53/0.8650, and 28.24/0.8739 for 0, 12, and 24 additional views, respectively. These results indicate that increasing the number of generated views improves performance. Note that these images cannot directly train 3DGS due to unknown camera poses. 

\section{Future Directions}
Future work may further enhance GlowGS by incorporating Vision-Language Models (VLMs) \cite{bai2025qwen3} and stronger Vision Foundation Models (VFMs) for robust semantic segmentation and depth estimation in adverse weather \cite{zhang2025mamba, yang2025erf, zhang2024heap, yan2025er, Yan_2025_CVPR}, visual reasoning \cite{du2026weatherreasonseg, ke2026view2space, Ke_2025_ICCV, ke2024hydra, ke2025explain, chen2025hypospace, chen2025auto}, geometric completion/correction \cite{lascomp, Yan_2023_ICCV}, and scene enhancement \cite{li2024nightcc, li2026bridging, chen2024dual}. In addition, integrating richer modalities, such as segmentation priors, polarization cues \cite{Yuan2025b, Yuan2025c, Yuan2025d}, and physics-based computational imaging constraints \cite{Gao2024, Yuan2025a, Zhang2025, Yuan2025e}, could provide complementary structural and physical information for more robust nighttime glow reconstruction and novel-view synthesis.

\section{Ethical Considerations} This paper introduces a new nighttime glow dataset, comprising 18 scenes. Each scene contains approximately 30 images, all affected by glow effects. To ensure privacy and ethical compliance, our collected images do not include any identifiable individuals, vehicles, or sensitive content. Additionally, our dataset is intended solely for research purposes, focusing on improving nighttime scene reconstruction without infringing on personal privacy or security concerns. We adhere to ethical data collection guidelines and ensure that no copyrighted or restricted content is included in our dataset.

\end{document}